\documentclass[letterpaper]{article} 
\usepackage{chapterbib}
\usepackage{cuted} 
\usepackage{multicol}
\usepackage{multirow}
\usepackage{booktabs} 
\usepackage[shortlabels]{enumitem}
\usepackage[table]{xcolor}
\usepackage{amsfonts}
\usepackage{amsmath} 

\usepackage{aaai2026}              

\usepackage{times}  
\usepackage{helvet}  
\usepackage{courier}  
\usepackage[hyphens]{url}  
\usepackage{graphicx} 
\usepackage{natbib}  
\usepackage{caption} 
\usepackage{algorithm}
\usepackage{algorithmic}
\usepackage{makecell}
\definecolor{mygray}{HTML}{f0f0f0}
\usepackage{newfloat}
\usepackage{listings}
\DeclareCaptionStyle{ruled}{labelfont=normalfont,labelsep=colon,strut=off} 
\floatstyle{ruled}
\newfloat{listing}{tb}{lst}{}
\floatname{listing}{Listing}

\title{Learning Procedural-aware Video Representations through State-Grounded Hierarchy Unfolding}
\author {
    Jinghan Zhao\textsuperscript{\rm 1},
    Yifei Huang\textsuperscript{\rm 2},
    Feng Lu\textsuperscript{\rm 1}\thanks{Corresponding author}
}
\affiliations {
    \textsuperscript{\rm 1}State Key Laboratory of VR Technology and Systems, School of CSE, Beihang University\\
    \textsuperscript{\rm 2}The University of Tokyo, Japan\\
    \{zhaojinghan,lufeng\}@buaa.edu.cn, hyf@iis.u-tokyo.ac.jp\\

} %

\usepackage{bibentry}

\usepackage[dvipsnames]{xcolor}
\usepackage{tcolorbox}
\tcbset{
  myprompt/.style={
    colback=gray!5!white,
    colframe=RoyalBlue,
    boxrule=0.8pt,
    left=6pt,right=6pt,top=6pt,bottom=6pt,
    fonttitle=\bfseries\color{RoyalBlue},
    title=Prompt
  }
}
\newcommand{\kw}[1]{\textcolor{OrangeRed!80!black}{\textbf{#1}}}
\newcommand{\meta}[1]{\textcolor{ForestGreen}{\texttt{#1}}}

\begin{document}
 
\maketitle

\begin{abstract}
Learning procedural-aware video representations is a key step towards building agents that can reason about and execute complex tasks. Existing methods typically address this problem by aligning visual content with textual descriptions at the task and step levels to inject procedural semantics into video representations. However, due to their high level of abstraction, `task' and `step' descriptions fail to form a robust alignment with the concrete, observable details in visual data. To address this, we introduce `states', \textit{i.e.}, textual snapshots of object configurations, as a visually-grounded semantic layer that anchors abstract procedures to what a model can actually see. We formalize this insight in a novel Task-Step-State (TSS) framework, where tasks are achieved via steps that drive transitions between observable states. To enforce this structure, we propose a progressive pre-training strategy that unfolds the TSS hierarchy, forcing the model to ground representations in states while associating them with steps and high-level tasks.
Extensive experiments on the COIN and CrossTask datasets show that our method outperforms baseline models on multiple downstream tasks, including task recognition, step recognition, and next step prediction. Ablation studies show that introducing state supervision is a key driver of performance gains across all tasks. Additionally, our progressive pretraining strategy proves more effective than standard joint training, as it better enforces the intended hierarchical structure.
\end{abstract}


\begin{links}
    \link{Code}{https://github.com/zhao-jinghan/TSS-unfolding}
\end{links}

\section{Introduction}

\begin{figure}[!htb]
\centering
\includegraphics[width=\linewidth, page=1]{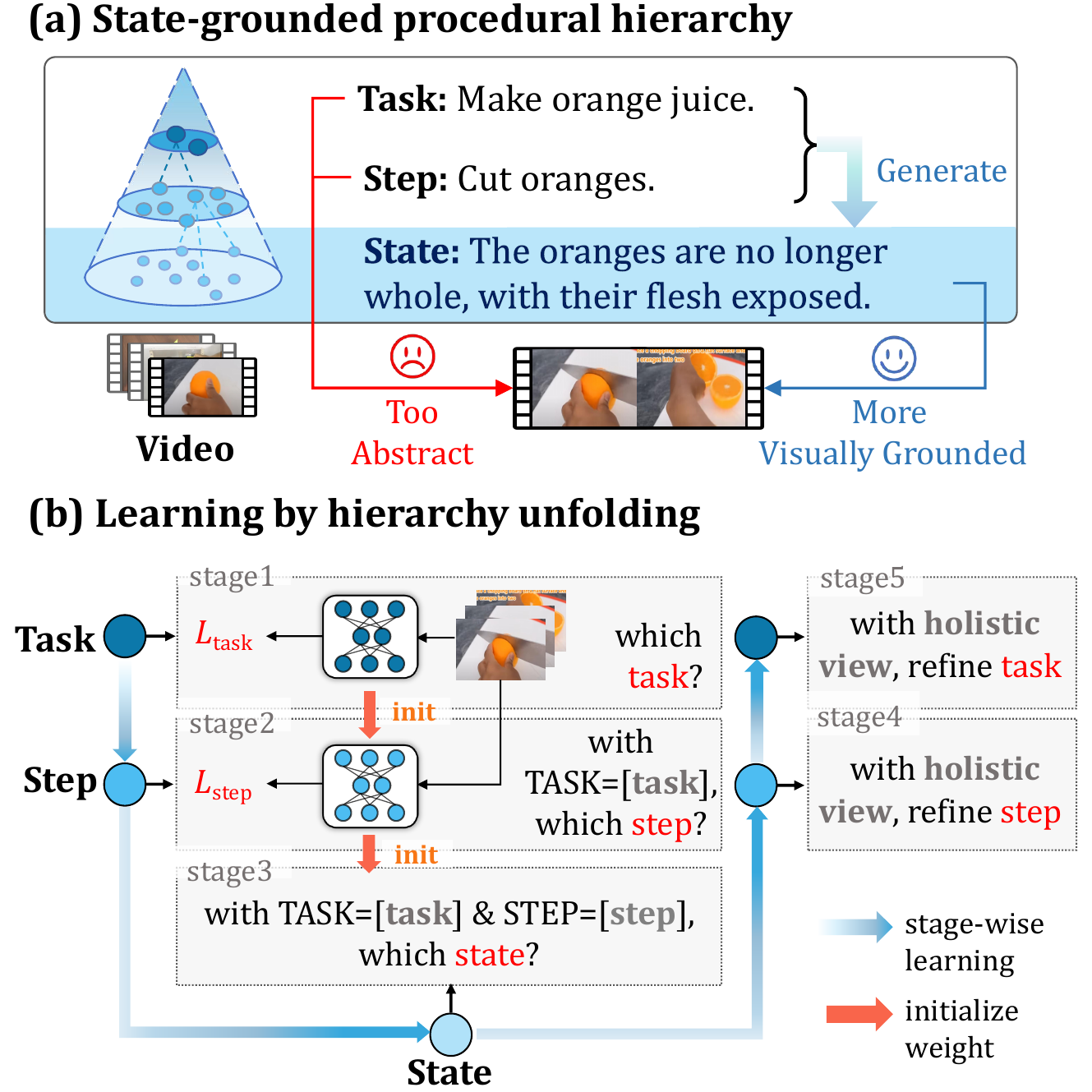}
\caption{(a) Our method extends the conventional task-step hierarchy by adding a state layer, forming the Task-Step-State (TSS) framework, where state can anchor abstract procedures to visual evidence. (b) To unfold the structure in TSS, we introduce a progressive pre-training strategy by cycling through the order: task→step→state→step→task.}
\label{fig1-TSS}
\end{figure}

Understanding and performing goal-oriented procedural activities is a core capability for intelligent agents. Humans effortlessly acquire such procedural knowledge from instructional videos, learning not only the overall goals but also the sequence of steps required to achieve them. Enabling machines to learn similar procedural-aware video representations is therefore central to progress in areas such as embodied AI, intelligent assistants, and emerging paradigms like cobodied intelligence (human-AI co-embodied intelligence) \cite{lu2025towards}.

To this end, existing methods typically seek to inject procedural knowledge into video representations by aligning visual content with textual descriptions from external knowledge bases \cite{lin2022learning,zhou2023procedure,mavroudi2023learning}. This alignment is generally performed at two levels of abstraction: the overall task (\textit{e.g.}, ``Make orange juice.") and the constituent steps (\textit{e.g.}, ``cut oranges"). By matching video features or ASR transcripts to these descriptions, models learn to recognize the high-level structure of an activity.

However, we argue that this two-level approach suffers from a critical limitation. Due to their high level of abstraction, ``task" and ``step" descriptions often fail to form a robust alignment with the concrete, observable details in visual data. A significant semantic gap  \cite{wang2023event} exists between an abstract command like ``cut oranges" and the raw pixels depicting the action. Thus, when training with visual-text alignment, this gap makes it difficult for models to ground abstract procedures in what they can actually see.

To address this, we introduce ``states", \textit{i.e.}, textual snapshots of object configurations, as a visually-grounded semantic layer that anchors abstract procedures to visual evidence. A state describes the observable configuration of key objects, such as ``the oranges are no longer whole, with their flesh exposed."

From a logical standpoint, states form the skeleton of any procedural task. A high-level ``Task" can be viewed as a macro-level transition from an initial state to a final state, while a series of ``Steps" are the actions that drive the transitions between these intermediate states. Building on this insight, we formalize this structure in a novel Task-Step-State (TSS) framework. In this hierarchy, tasks are achieved via steps, and steps are semantically anchored by the observable state transitions they cause.

To enforce this structure and ensure the model learns the intended hierarchy, we propose a novel progressive pre-training strategy that unfolds the TSS hierarchy. Instead of learning all semantic levels jointly, our strategy forces the model to ground its representations in the most concrete layer—the states. By this means, it better learns to associate states with steps, and finally, steps with high-level tasks. This progressive approach ensures that the learning of abstract concepts is firmly built upon a foundation of visually-grounded understanding.

We validate our approach through extensive experiments on the COIN \cite{tang2019coin} and CrossTask \cite{zhukov2019cross} datasets. Our method outperforms baseline models on multiple downstream tasks, including task recognition, step recognition, and next step prediction. Furthermore, detailed ablation studies confirm our core hypotheses: introducing state supervision is the key driver of these performance gains, and our progressive pre-training strategy is significantly more effective than standard joint training at enforcing the hierarchical structure. Our contributions are:
\begin{itemize}
    \item We propose a novel Task-Step-State (TSS) framework that introduces ``states" as a visually-grounded semantic layer, effectively bridging the semantic gap between abstract procedural texts and concrete visual content.
    \item We introduce a progressive pre-training strategy that unfolds the TSS hierarchy, ensuring that representations are grounded in concrete states before learning abstract steps and tasks, better enforcing the procedural logic.
    \item Experiments demonstrate that our method achieves consistent performance gains over existing models and validate that both the TSS framework and the progressive training strategy are key to this success.

\end{itemize}

\section{Related works}

\subsection{Procedural Learning from Instructional Video}

Instructional videos serve as a rich source for learning how agents perceive and act in the world, fueling applications such as procedure planning \cite{wang2023pdpp,yang2025planllm,ilaslan2025vg,wu2024open}, step recognition \cite{kazakos2021little,piergiovanni2021unsupervised,zhukov2019cross}, localization \cite{dvornik2022graph2vid}, and task recognition \cite{ghoddoosian2022hierarchical}. The introduction of large-scale, richly annotated datasets \cite{zhou2018towards,tang2019coin,afouras2023ht} has greatly advanced this field. However, acquiring dense temporal annotations for task-step structures remains a key challenge, limiting the scalability of supervised approaches.

To address this bottleneck, recent efforts have embraced weakly-supervised paradigms that leverage external knowledge sources, such as wikiHow \cite{koupaee2018wikihow}, to generate supervisory signals for large-scale, unannotated video collections \cite{lin2022learning,zhou2023procedure,mavroudi2023learning,samel2025leveraging,zhong2023learning}. These approaches typically aim to learn procedural-aware video representations by aligning visual or transcribed-text features with textual descriptions of steps. For instance, Paprika \cite{zhou2023procedure} utilizes pre-trained visual and textual encoders to perform cross-modal matching, and builds a graph that captures the temporal progression of both visual content and procedural steps derived from wikiHow. Similarly, \cite{mavroudi2023learning} integrates video frames, ASR-generated narrations, and step descriptions, proposing dual alignment paths (``step–frame" and ``step–narration–frame") to bridge the gap between vision and language modalities.

While effective at capturing the high-level procedure, these approaches operate purely at the abstract task and step levels. This introduces a critical limitation, as noted by works like \cite{wang2023event}: a semantic gap exists between abstract, action-centric text and concrete visual data. Our work directly confronts this gap. We argue that to truly ground procedural learning, the knowledge hierarchy itself must be augmented. We achieve this by introducing ``states" as a new, visually-grounded semantic layer, bridging the gap between abstract commands and observable visual content.

\subsection{State-Based/Object-Centric Video Understanding}

The concept of modeling object states and their transitions is a cornerstone of robust action understanding
\cite{aboubakr2019recognizing,saini2022recognizing,doughty2020action}, procedural planning \cite{wang2023pdpp,zhao2022p3iv,ilaslan2025vg,wu2024open}, and beyond.
Recent work has focused on automatically extracting and utilizing state information from video \cite{alayrac2017joint,souceklook,souvcek2022multi,xue2024learning,niu2024schema}. For instance, \cite{souvcek2024genhowto} mines triplets of (initial state, action, end state) from instructional videos to train generative models of object states.
\cite{niu2024schema} represent each step as a change of state in procedural planning, supplement state descriptions via Large Language Model (LLM), and align them with visual observations to achieve tracking of state changes.

Our work builds directly on these insights. We adopt the effective strategy of using LLMs to generate state descriptions, but move beyond simply tracking them. We are the first to formalize states as a fundamental component within a three-tiered hierarchy (Task-Step-State), specifically for pre-training procedural-aware video representations.

\subsection{Progressive and Curriculum-Based Learning}
Progressive learning draws inspiration from the way humans acquire skills, where knowledge is built up gradually in a meaningful order. In machine learning, this idea has been formalized primarily through two paradigms: curriculum learning, which trains models from easy to hard examples \cite{bengio2009curriculum,hacohen2019power}, and hard example mining, which does the reverse by focusing on difficult samples first \cite{sun2018glancenets,suh2019stochastic}.
In recent years, progressive learning has been applied in various domains from image generation to gait recognition \cite{du2021progressive,wang2018fully,dou2022gaitmpl,garg2025stpro}.
Recent work in video understanding has also adopted this principle. For instance, STPro \cite{garg2025stpro} uses a spatio-temporal curriculum, progressively increasing the complexity of action sequences and scene backgrounds to improve grounding.

Our work introduces a novel application of this progressive philosophy. Our proposed pre-training pathway enables the model to unfold structured procedural knowledge within our TSS framework, leading to more robust and generalizable procedural-aware video representations.

\section{Method}

Our method is designed around two core contributions to learn procedural-aware video representations. We first introduce the Task-Step-State (TSS) framework, a new knowledge structure that grounds abstract procedures in observable visual content. We then propose a novel progressive pre-training strategy that teaches a model to understand this hierarchy by unfolding it in a principled, stage-wise manner.

\subsection{Problem Formulation}


We aim to learn robust video representations guided by structured procedural knowledge. Starting from a textual corpus, such as wikiHow \cite{koupaee2018wikihow}, consisting of tasks decomposed into sequential steps, our goal is to train a video encoder $\Phi$ mapping video clips into representations useful for procedural understanding tasks (e.g., task recognition, step recognition, and forecasting).

Formally, given video clips $v$, the encoder $\Phi$ produces representations: $f_v=\Phi(v)$. Representation quality is assessed through downstream task performance.

\subsection{TSS Framework Construction}
\begin{figure}[!h]
\centering
\includegraphics[width=\linewidth, page=1]{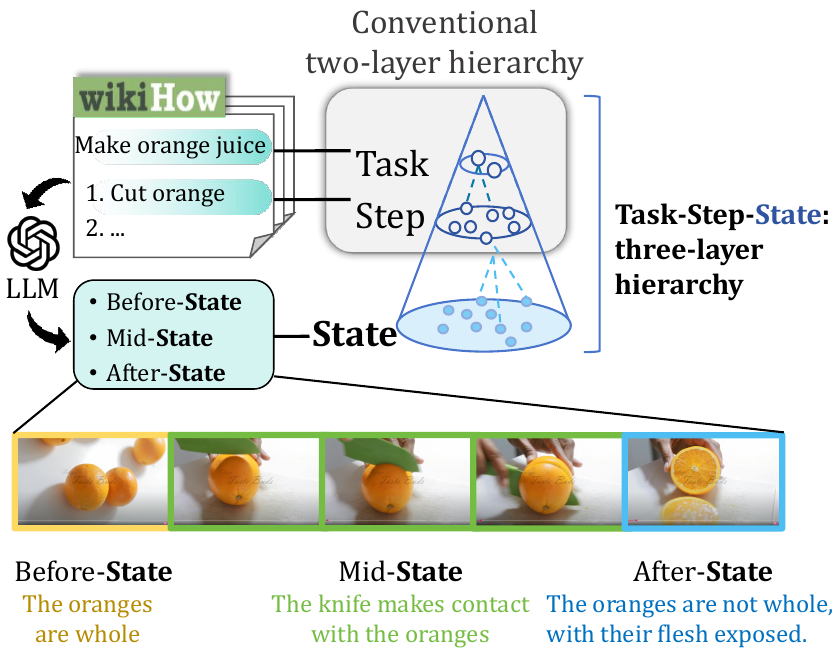}
\caption{The TSS framework extends the task-step hierarchy with an LLM-generated state layer.}
\label{fig-TSSgen}
\end{figure}

The foundation of our approach is the creation of a new, state-augmented knowledge framework to bridge the semantic gap between abstract steps and concrete visual content. 
\subsubsection{Grounding Procedures in States}
The conventional structure for procedural knowledge, used in prior work, is a two-level hierarchy. A high-level \texttt{task} (denoted by $T$) is decomposed into an ordered sequence of \texttt{steps} (denoted by $s$). However, as we have established, the purely abstract nature of these descriptions is the source of the critical semantic gap that hampers robust alignment with visual data.
To resolve this, we augment the structure by introducing a third, more visually grounded layer: the \texttt{state}, denoted by $c$. We define a state as a \textit{textual snapshot of object configurations}. This includes not only the properties of single objects, such as their attributes, quantity, and shape, but also the spatial relationships between them. To capture the temporal nature of actions, denote the $j-th$ step of the $i-th$ task as $s_{i,j}$, we associate each $s_{i,j}$ from the original corpus with three distinct state types: a before-state ($c_{i,j}^b$), a mid-state ($c_{i,j}^m$), and an after-state ($c_{i,j}^a$). This enriches each step into an expanded tuple: $$s'_{i,j} = (s_{i,j}, c_{i,j}^b, c_{i,j}^m, c_{i,j}^a),$$ forming our proprietary three-level hierarchy.

Specifically, we construct our state-augmented knowledge base from the wikiHow corpus. As manual annotation of states for every step is infeasible, we employ a Large Language Model (LLM) GPT-4o-mini to automatically infer and generate the state descriptions, which is a practical approach with precedents in related work \cite{niu2024schema}. For each step description $s_{i,j}$, the LLM is prompted to generate the corresponding before-state, mid-state, and after-state. To ensure the logical consistency and quality of the generated text, we utilize a Chain-of-Thought (CoT) \cite{wei2022chain} prompting strategy. This automated process effectively transforms the standard wikiHow corpus into our rich, state-aware TSS framework, which we will then use for the next step: pseudo-label generation.

\begin{figure*}[!t]
\centering
\includegraphics[width=\linewidth, page=1]{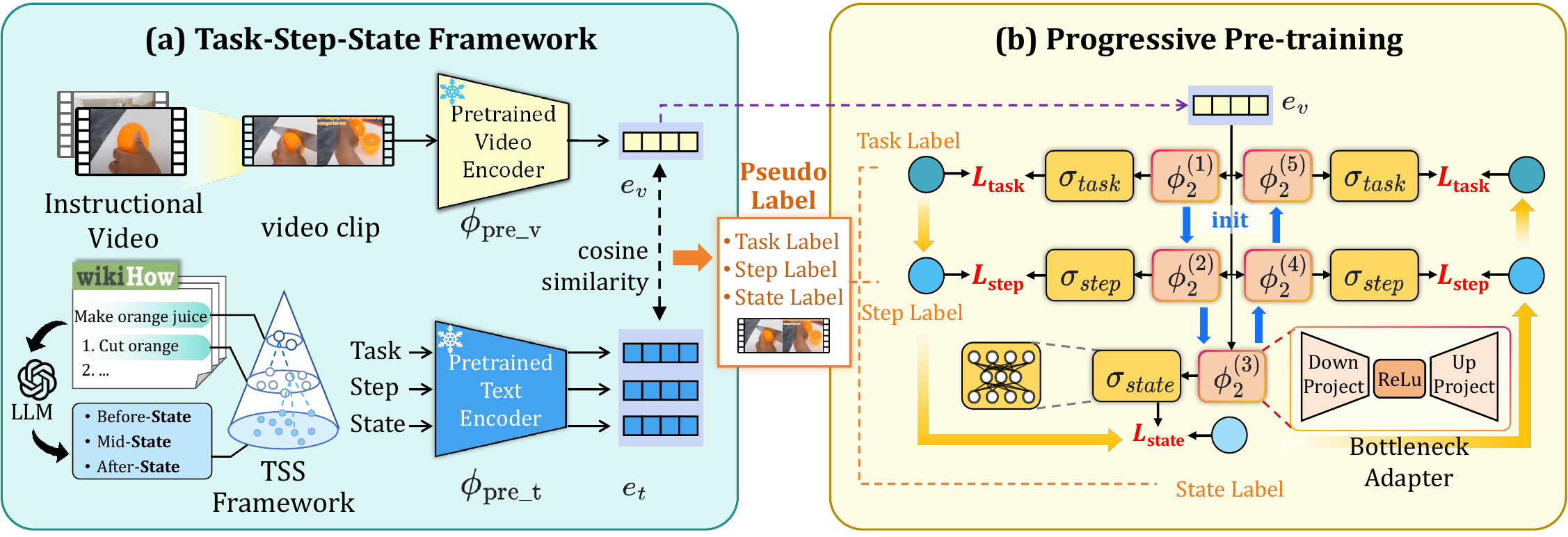}
\caption{Overview.
(a) First, an LLM generates state descriptions from the task and step in WikiHow, yielding the three-level TSS framework. We encode these texts with a frozen text encoder $\phi_{\text{pre\_t}}$ and encode video clips from HowTo100M with a frozen vision encoder$\phi_{\text{pre\_v}}$, producing text features $e_t$ and video features $e_v$. Cosine similarity between $e_t$ and $e_v$ creates pseudo labels at each level. (b) Second, the progressive pre-training strategy trains the vision model in five stages, each focused on one TSS layer. Every stage takes $e_v$, passes it through an adapter $\phi_2^{(i)}$ and a task-specific head $\sigma$, and is supervised by the corresponding pseudo label.}
\label{fig-main}
\end{figure*}


\subsubsection{Pseudo-Label Generation from TSS} To generate supervisory signals for our pre-training, we align a large corpus of unannotated videos with our Task-Step-State (TSS) framework. This process creates a rich set of matched video segments and the textual descriptions of tasks, steps, and states, which can be further used in pre-training. 

We begin by projecting video data and text descriptions into a shared embedding space. Using off-the-shelf pre-trained vision and text encoders $\phi_{\text{pre\_v}}(\cdot)$ and $\phi_{\text{pre\_t}}(\cdot)$, we obtain initial embeddings $e_v=\phi_{\text{pre\_v}}(v)$ and $e_t=\phi_{\text{pre\_t}}(t)$ for each video clip $v$ and text description $t$. The correspondence between these initial embeddings is measured by their cosine similarity:
\begin{equation}
S(e_v, e_t) = \frac{e_v \cdot e_t}{|e_v| |e_t|}
\label{eq:cosine_similarity}
\end{equation}

To create a stable and canonical set of targets, we then cluster the text embeddings $e_t$ at each semantic level. For instance, all step description embeddings are clustered to form a set of representative ``step nodes", $\mathcal{N}_{\text{step}}$. This process is repeated for tasks ($\mathcal{N}_{\text{task}}$) and for each type of the before/mid/after state ($\mathcal{N}_{\text{state}}$) separately, to yield $\mathcal{N}^b_{\text{state}}$, $\mathcal{N}^m_{\text{state}}$ and $\mathcal{N}^a_{\text{state}}$.

With these semantic nodes established, we generate pseudo-labels by aligning each video clip embedding $e_v$ to them. We calculate the similarity of $e_v$ to all nodes within a given set (\textit{e.g.}, $\mathcal{N}_{\text{step}}$) and select the top-k most similar nodes as positive targets for a multi-class classification objective. This yields our primary supervision labels of Video-Node Matching: \begin{itemize}
    \item \textbf{\texttt{taskVNM}}: Generated by matching video features against task nodes ($\mathcal{N}_{\text{task}}$).
    \item \textbf{\texttt{stepVNM}}: Generated by matching video features against step nodes ($\mathcal{N}_{\text{step}}$).
    \item \textbf{\texttt{stateVNM}}: First, the node set ($\mathcal{N}_{\text{state}}^b$/$\mathcal{N}_{\text{state}}^m$/$\mathcal{N}_{\text{state}}^a$) is determined by the best-matching node, and then the top-k matching nodes are retrieved.
    
\end{itemize}

Finally, inspired by prior work on learning procedural graphs \cite{zhou2023procedure}, we extract two additional supervision signals to capture richer step-level relationships. By analyzing the graph structure formed from step nodes, we derive:
\begin{itemize}
    \item \textbf{\texttt{stepNRL}} (Node Relation Learning) : trains the model to identify a step's preceding and succeeding nodes.
    \item \textbf{\texttt{stepTCL}} (Task Context Learning) : trains the model to recognize other step nodes belonging to the same overarching task.
\end{itemize}

\subsection{Progressive Pre-Training}
One critical aspect of our method lies in how we teach the model the TSS hierarchy. Instead of a black-box joint training approach, we propose a principled, progressive pre-training strategy that unfolds the hierarchy in a specific, highly effective pathway.

\subsubsection{Model Architecture and Preliminaries}
Our model, denoted as $\Phi$, uses a parameter-efficient design to make large-scale pretraining computationally affordable. It is composed of two key components:
\begin{itemize}

    \item \textbf{A frozen pre-trained vision encoder, $\phi_{\text{v}}$. }In our work, we use vision encoder of S3D model \cite{miech2020end} simultaneously as $\phi_{\text{v}}$ and $\phi_{\text{pre\_v}}$, keeping its weights frozen throughout our training.

    \item \textbf{A trainable lightweight adapter $\phi_2$.} The adapter features a bottleneck architecture \cite{houlsby2019parameter,sung2022vl}, consisting of a down-projection layer followed by an up-projection layer. This lightweight module is the primary component we train, allowing it to adapt the generic visual features from $\phi_{\text{v}}$ for our specific procedural understanding tasks.
\end{itemize}

\noindent The final video representation $f_v$ is thus computed as: $$f_v = \Phi(v) = \phi_2(\phi_{\text{v}}(v))=\phi_2(e_v)$$ 
This representation is then fed into various \textbf{task-specific heads}, denoted by $\sigma(\cdot)$, which are simple Multi-Layer Perceptrons (MLPs) designed for multi-class classification objectives defined by our pseudo-labels.

\subsubsection{Progressive Pre-training Strategy}

With this architecture in place, we address a fundamental question: \textit{how can the intrinsic, hierarchical connections between the Task, Step, and State layers be best exploited for representation learning?} A standard approach of jointly training on all supervisory signals at once treats the learning process as a black box, offering little insight into how these distinct knowledge types influence one another.

To answer this question, we move beyond joint training and propose a \textbf{progressive pre-training methodology}. Our work extends this concept to explore the optimal learning pathway for a hierarchical knowledge system like TSS. Our work asks more complex questions: Does knowledge transfer exclusively downward (\textit{e.g.}, from Task to Step), or can knowledge from a lower level (like State) be used to retrospectively refine an understanding of a higher level (like Step or even Task)?

To systematically explore these questions, we define a training process that proceeds in a sequence of stages. This approach allows us to control the flow of information and investigate the impact of different learning pathways. The mechanism is as follows:
\begin{itemize}
    \item \textbf{Staged Learning:} The overall training is decomposed into a series of stages. Each stage $i$ is designed to learn a specific procedural knowledge, and corresponds to different task-specific heads($\sigma_{task}$/$\sigma_{step}$/$\sigma_{state}$).

    \item \textbf{Knowledge Propagation via Adapter:} In each stage, we train the adapter ($\phi_2^{(i)}$) alongside a set of randomly initialized task heads required for that stage's objectives. Upon completion, we discard the task heads but preserve the weights of the trained adapter to initialize the adapter($\phi_2^{(i+1)}$) for the subsequent stage ($i+1$). The task heads for the new stage are again initialized randomly.
\end{itemize}

This progressive process, where knowledge is accumulated and passed through the adapter, allows us to assess different learning trajectories.

Through systematic exploration, we design our primary learning strategy around a full cycle: \textbf{Task$\rightarrow$Step $\rightarrow$State$\rightarrow$Step$\rightarrow$Task}. This pathway is designed to model a complete analysis-synthesis process. It first performs top-down analysis, using high-level context to predict fine-grained details, before performing bottom-up synthesis, using grounded details to verify and confirm high-level hypotheses. This circular pathway enables bidirectional information flow, for instance, allowing state-level knowledge to refine step representations. Furthermore, each segment of this cycle can be treated as an independent progressive strategy, enabling us to ablate and evaluate the contribution of each knowledge transfer direction in our experiments.

\section{Experiments}
We conduct a comprehensive set of experiments designed to first dissect our proposed method and then validate its superiority. We begin with detailed ablation studies to answer two fundamental questions: 
\textbf{(1)} What is the optimal progressive pathway for learning the Task-Step-State hierarchy? \textbf{(2)} How does each semantic layer contribute to the final representation? Having identified the best-performing configuration from these studies, we then \textbf{(3)} compare our final model against state-of-the-art methods to demonstrate its effectiveness on standard procedural understanding benchmarks.

\subsection{Datasets}
For pre-training, our visual data is an 85K video subset of HowTo100M \cite{miech2019howto100m}, following \cite{bertasius2021space}. Our textual data is a curated subset of WikiHow \cite{koupaee2018wikihow} comprising 1,053 tasks and 10,588 steps. 
We evaluate our representations on two standard benchmarks: COIN \cite{tang2019coin} and CrossTask \cite{zhukov2019cross}. COIN consists of 11.8K videos across 180 tasks. CrossTask includes 83 tasks, of which only 18 have full temporal annotations.

\subsection{Evaluation Protocol}
Consistent with prior works \cite{zhou2023procedure,lin2022learning}, we assess the quality of our learned representation on three downstream tasks: \textbf{Task Recognition (TR)}, \textbf{Step Recognition (SR)}, and \textbf{Step Forecasting (SF)}.  We report Top-1 Accuracy as the primary metric. For fine-tuning, we attach two distinct evaluation models to our frozen representation: a simple MLP and a more powerful one-layer Transformer \cite{vaswani2017attention}.
This allows us to assess the robustness of our learned features across downstream models of varying capacities. The specific architectural details for both evaluation models are deferred to the supplementary material in our GitHub repository.

\subsection{Implemental Details}

For data processing, we segment the source videos into non-overlapping 9.6-second clips. For each of the 10,588 procedural steps, we generate three types of state descriptions, resulting in a total of 31,764 state-level texts. The video feature $e_v$ and text feature $e_t$ both have a dimensionality of 512. Our adapter employs a bottleneck architecture, projecting the 512-dimensional features down to a hidden dimension of 128 before restoring them to 512.
We train our model with a batch size of 256 using the Adam optimizer. All experiments were conducted on a single server equipped with eight H200 GPUs. Further details regarding hyperparameters and training procedures are provided in our GitHub repository.

\subsection{Pathway and Component Analysis}
We first conduct an extensive analysis to understand the impact of our core design choices, focusing on the pre-training curriculum.

\begin{table}[h]
\centering
\setlength{\tabcolsep}{12pt}        
\begin{tabular}{l|l}
\Xhline{0.8pt}                     
\bf ID       & \bf Pre-training Pathway \\ \hline
Path-1      & Task \\ 
Path-2      & Task$\rightarrow$Step \\ 
Path-3 (Ours)      & Task$\rightarrow$Step$\rightarrow$\textit{State} \\ 
Path-4 (Ours)      & Task$\rightarrow$Step$\rightarrow$\textit{State}$\rightarrow$Task \\ 
Path-5 (Ours)      & Task$\rightarrow$Step$\rightarrow$\textit{State}$\rightarrow$Step \\ 
Path-6 (Ours)      & Task$\rightarrow$Step$\rightarrow$\textit{State}$\rightarrow$Step$\rightarrow$Task \\ 

\Xhline{0.8pt}                   
\end{tabular}
\caption{Proposed progressive pre-training pathways}
\label{table-pathway}
\end{table}

\begin{table*}[t]
\centering
\begin{tabular}{l| ccc| ccc| ccc |ccc}
\multicolumn{1}{l}{} &
\multicolumn{6}{c}{\textbf{Downstream MLP}} &
\multicolumn{6}{c}{\textbf{Downstream Transformer}} \\
\Xhline{1pt}
 &
\multicolumn{3}{c|}{\textbf{COIN}} & \multicolumn{3}{c|}{\textbf{CrossTask}} &
\multicolumn{3}{c|}{\textbf{COIN}} & \multicolumn{3}{c}{\textbf{CrossTask}} \\

\multirow{-2}{*}{\textbf{Pretrain Pathway}} &
\textit{TR} & \textit{SR} & \textit{SF} &
\textit{TR} & \textit{SR} & \textit{SF} &
\textit{TR} & \textit{SR} & \textit{SF} &
\textit{TR} & \textit{SR} & \textit{SF} \\
\Xhline{0.6pt}
No pretrain & 2.09 & 1.37 & 0.84 & 53.00 & 20.21 & 23.27 & 78.31 & 39.23 & 35.43 & 89.44 & 56.82 & 56.17 \\
\Xhline{0.6pt}
Path-1                               & 73.31 & 34.18 & 23.67 & 89.44 & 52.09 & 52.40 & 82.25 & 37.84 & 37.70 & 89.03 & 54.58 & 53.67 \\
Path-2                          & 82.45 & 43.06 & 36.04 & 89.44 & 56.55 & 56.37 & 83.30 & 40.96 & 40.03 & 90.06 & 55.71 & 55.83 \\
Path-3 (Ours)                   & 80.73 & 41.28 & 34.35 & 89.65 & 55.16 & 55.27 & 82.16 & 40.24 & 39.22 & 88.82 & 54.84 & 54.13 \\
Path-4 (Ours)                & 77.74 & 37.51 & 24.84 & 89.86 & 54.29 & 54.57 & 81.16 & 36.04 & 36.85 & 89.65 & 55.34 & 54.20 \\
Path-5 (Ours)               & 83.78 & 44.54 & 38.07 & 89.44 & 57.92 & 57.13 & 83.11 & 42.42 & 40.40 & 89.44 & 57.08 & 56.50 \\
Path-6 (Ours)          & 83.30 & 44.04 & 36.94 & 90.27 & 57.63 & 57.13 & 83.25 & 42.52 & 40.02 & 89.44 & 57.08 & 56.50 \\

\Xhline{0.6pt}
Mix-Train                          & 77.74 & 38.43 & 29.79 & 89.23 & 53.02 & 53.17 & 82.92 & 37.93 & 38.13 & 89.03 & 54.06 & 53.70 \\
Fusion-Concat                      & 83.16 & 44.21 & 36.65 & 89.65 & 57.34 & 57.27 & 82.49 & 42.34 & 40.10 & 90.27 & 56.70 & 56.17 \\
Fusion-AvgPool                     & 83.11 & 44.35 & 36.88 & 90.06 & 57.48 & 57.50 & 83.40 & 42.50 & 40.20 & 90.68 & 56.84 & 56.10 \\
\Xhline{1pt}
\end{tabular}
\caption{Ablation results. (1) We compare the downstream accuracy of six progressive pre-training pathways and joint training. (2) We fuse video representations pre-trained with three different levels and evaluate two fusion strategies.}
\label{table-ablation}
\end{table*}

\subsubsection{Analysis of the Pre-training Pathway}
To investigate the optimal pre-training curriculum for incorporating hierarchical procedural knowledge, we first experiment with six distinct pre-training paths and compare their performance on downstream tasks. The sequential paths investigated are listed in Table \ref{table-pathway}.

\begin{figure}[ht]
\centering
\includegraphics[width=0.95\linewidth, page=1]{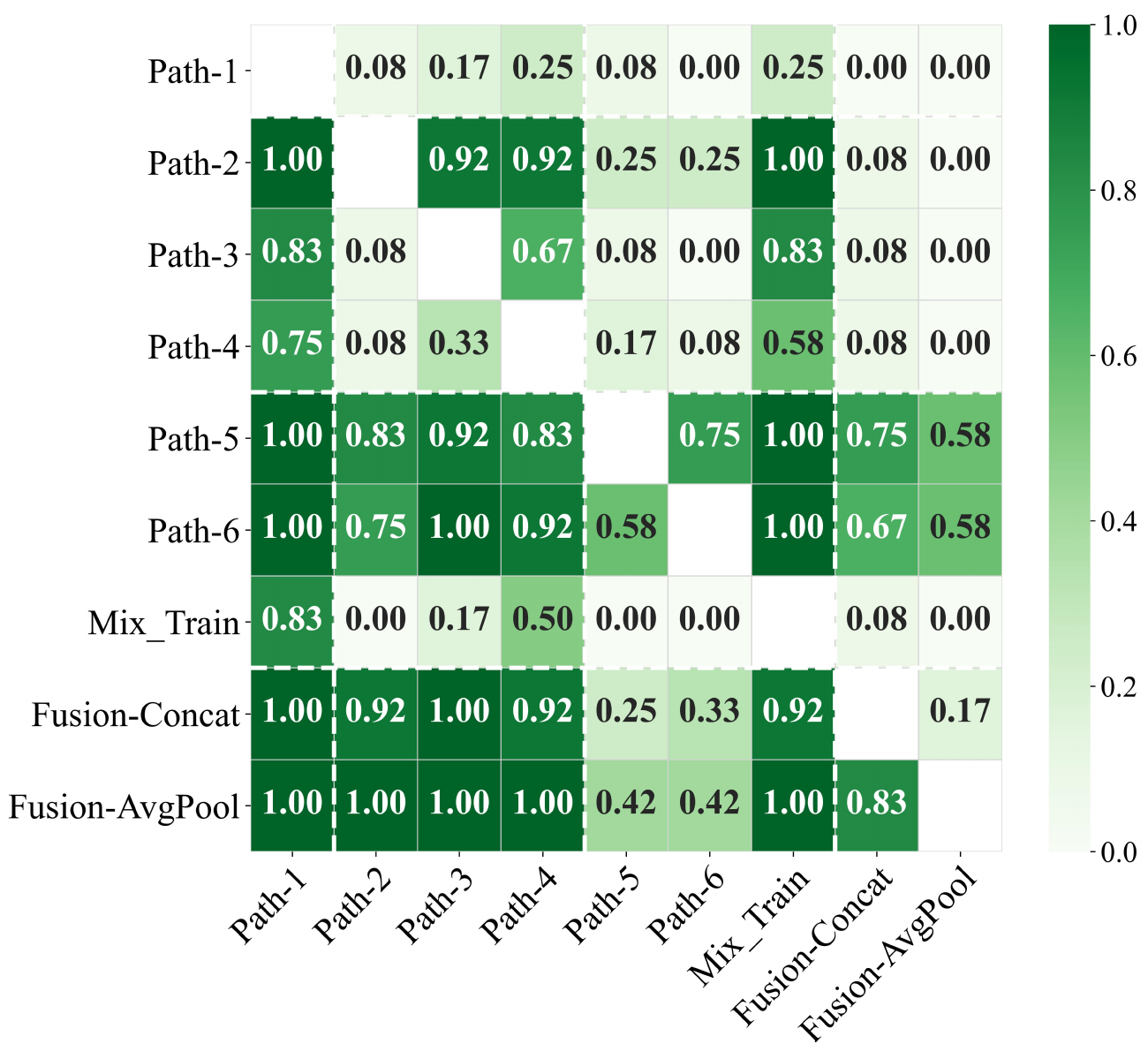}
\caption{Ablation results Comparison. The value at position $(i, j)$ in the heat map denotes the fraction of 12 evaluation settings where configuration $i$ outperforms configuration $j$.}
\label{fig4-ablation}
\end{figure}

The results are presented in Table \ref{table-ablation}. Additionally, we report the results of MIL-NCE \cite{miech2020end} to serve as our ``No pretrain" baseline for comparison. MIL-NCE is a pre-training objective for video-narration contrastive learning. For this baseline, we employ the frozen S3D model provided by the authors to extract video features. We visualized the performance matrix for all ablation settings as a heatmap (Figure \ref{fig4-ablation}) for better visibility.

From Table~\ref{table-ablation}, it is clear that pre-training on task and step information is beneficial, with Path-2 outperforming Path-1 and the ``No Pre-train" baseline. This aligns with prior work. More insights emerge when we introduce the state layer.



Path-3 (Task $\rightarrow$ Step $\rightarrow$ State) is our most direct state-aware curriculum. This pathway reveals a crucial challenge: while grounding the representation in states is the goal, doing so naively causes a slight performance dip on the more abstract task and step recognition tasks. This happens because the model's focus shifts heavily towards the concrete vision-state alignment, temporarily de-emphasizing the higher-level semantics. This finding validates our core premise: simply adding states is not enough; how they are integrated is critical.

This is further evidenced by Path-4 (... $\rightarrow$ State $\rightarrow$ Task). Transitioning directly from the concrete state layer back to the highly abstract task layer results in comparable or even lower performance compared to Path-3 across most settings (\textit{e.g.}, TR of 81.16 on COIN). This observation powerfully supports the architectural rationale of our three-layer hierarchy, confirming that a significant semantic gap exists between states and tasks, and that the step layer serves as an essential intermediate bridge.



Our U-shaped pathway is designed to correctly utilize this bridge. Path-5 (... $\rightarrow$ State $\rightarrow$ Step) demonstrates the power of this analysis-synthesis approach. By using the now-grounded state knowledge to retrospectively refine its understanding of step semantics, it achieves top-tier performance, with Step Recognition (SR) and Step Forecasting (SF) scores of 42.42 and 40.40 on COIN, respectively.
This result reveals two key insights:
\begin{itemize}
    \item First, it confirms that state-level pre-training forces the model to anchor its understanding in observable evidence, which serves as a robust foundation for interpreting more abstract concepts.
    \item Second, it validates that our progressive pre-training is effective for transferring this visual grounding upward to enrich and refine the abstract step representations.
\end{itemize}

Path-6 extends Path-5 by incorporating the final Step $\rightarrow$ Task traceback. This addition only leads to marginal performance gains. We hypothesize that this is due to convergence, as the step representations may have already been sufficiently optimized in Path-5, thus limiting the marginal gain from the final, high-level objective.

\begin{figure}[t]
\centering
\includegraphics[width=\linewidth, page=1]{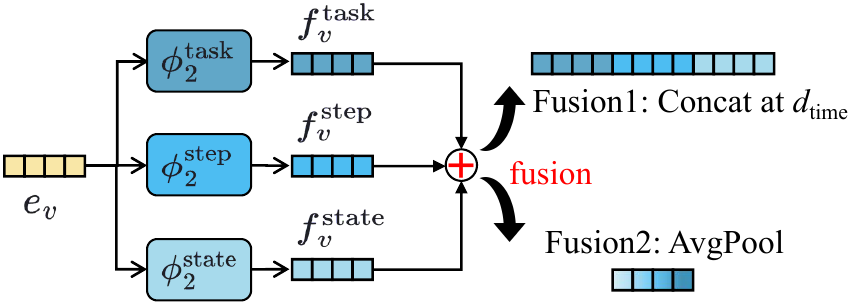}
\caption{Different ways to fuse the three visual representations after adapter fine-tuning in the ablation study. 
}
\label{fig3-fusiontype}
\end{figure}

\begin{table*}[t]
\centering

\setlength{\tabcolsep}{1.4mm}
\begin{tabular}{ll| ccc| ccc| ccc |ccc}

\multicolumn{2}{l}{} &
\multicolumn{6}{c}{\textbf{Downstream MLP}} &
\multicolumn{6}{c}{\textbf{Downstream Transformer}} \\
\Xhline{1pt}
 & &
\multicolumn{3}{c|}{\textbf{COIN}} & \multicolumn{3}{c|}{\textbf{CrossTask}} &
\multicolumn{3}{c|}{\textbf{COIN}} & \multicolumn{3}{c}{\textbf{CrossTask}} \\

\multicolumn{2}{c|}{\multirow{-2}{*}{\textbf{Pretrain Method}}} &
\textit{TR} & \textit{SR} & \textit{SF} &
\textit{TR} & \textit{SR} & \textit{SF} &
\textit{TR} & \textit{SR} & \textit{SF} &
\textit{TR} & \textit{SR} & \textit{SF} \\
\Xhline{0.6pt}
\multicolumn{2}{c|}{MIL-NCE \cite{miech2020end}} & 2.09 & 1.37 & 0.84 & 53.00 & 20.21 & 23.27 & 78.31 & 39.23 & 35.43 & 89.44 & 56.82 & 56.17 \\
\multicolumn{2}{c|}{Paprika \cite{zhou2023procedure}} & 81.54 & 42.39 & 34.10 & 89.65 & 56.21 & 55.77 & 82.83 & 41.19 & 38.93 & 90.27 & 55.57 & 55.67 \\
\Xhline{0.6pt}
\multirow{4}{*}{\textbf{Ours}}
 & Path-5     & \textbf{83.78} & \textbf{44.54} & \textbf{38.07} & 89.44 & \textbf{57.92} & \textbf{57.13} & \textbf{83.11} & \textbf{42.42} & \textbf{40.40} & 89.44 & \textbf{57.08} & \textbf{56.50} \\
 & Gains to Paprika & \underline{+2.24} & \underline{+2.15} & \underline{+3.97} & \underline{-0.21} & \underline{+1.71} & \underline{+1.36} & \underline{+0.28} & \underline{+1.23} & \underline{+1.47} & \underline{-0.83} & \underline{+1.51} & \underline{+0.83} \\
 & Path-6    & \textbf{83.30} & \textbf{44.04} & \textbf{36.94} & \textbf{90.27} & \textbf{57.63} & \textbf{57.13} & \textbf{83.25} & \textbf{42.52} & \textbf{40.02} & 89.44 & \textbf{57.08} & \textbf{56.50} \\
 & Gains to Paprika & \underline{+1.76} & \underline{+1.65} & \underline{+2.84} & \underline{+0.62} & \underline{+1.42} & \underline{+1.36} & \underline{+0.42} & \underline{+1.33} & \underline{+1.09} & \underline{-0.83} & \underline{+1.51} & \underline{+0.83} \\
\Xhline{1pt}
\end{tabular}
\caption{Accuracy of our method and the SOTA baselines on downstream procedural understanding tasks. Our method surpasses Paprika on all three tasks.}
\label{table-withSota}
\end{table*}

\subsubsection{Analysis of Alternative Learning Strategies}
Having established that our TSS framework progressively injects procedure knowledge to strengthen video representations, we next explore: can the same knowledge be fully exploited (1) without the staged curriculum, or (2) by directly fusing representations learned at different semantic levels? To answer this, we conduct two ablation studies in Table~\ref{table-ablation}:
\begin{itemize}
    \item Joint training: we investigate an alternative training strategy: joint pre-training (referred to as ``\textbf{Mix\_Train}"). In this setup, the model is supervised simultaneously by all objectives from the task, step, and state levels, rather than progressively.
    \item Multi-Level Feature Fusion: We try to integrate features learned from different stages of our progressive pre-training approach. Specifically, we investigate whether the three video representations obtained through distinct pre-training pathways ($f_v^{task}$, $f_v^{step}$, and $f_v^{state}$, derived from Path-1, Path-2, and Path-3, respectively) can be fused to achieve improved performance. We evaluated two fusion methods (Figure \ref{fig3-fusiontype}):
    (1) concatenation along the temporal dimension, and (2) average pooling.
\end{itemize}

For Mix\_Train strategy, its accuracy falls between Path-1 and Path-2. This result demonstrates that simply training on all objectives altogether is not optimal. This suggests that the progressive nature of our pathway is crucial, as it allows the model to effectively leverage the structural and causal relationships between the three semantic levels. The direct joint training approach fails to capture these relationships.

Regarding the two fusion approaches, the Fusion-AvgPool model is very strong. This strong performance provides further validation for our central thesis, demonstrating that the state layer contributes critical, complementary information that is missing from the standard Task-Step hierarchy. Meanwhile, our progressive model, Path-5, is still superior, particularly with downstream MLP. This proves that while adding state information is crucial, our progressive pre-training strategy is a more effective method for integrating this knowledge into a single, cohesive representation.

\subsubsection{Comparison to the State of the Art}
Having validated our TSS framework and identified Path-5 and 6 as our champion models, we now compare it against state-of-the-art (SOTA) baselines. As shown in Table~\ref{table-withSota}, for a fair comparison, all methods share the same frozen S3D encoder and downstream model architectures. Our primary SOTA baseline is Paprika \cite{zhou2023procedure}, which also leverages wikiHow but operates only at the step level.

Results are reported in Table \ref{table-withSota}, where our method consistently outperforms the state of the art. On the COIN dataset with a Transformer head, our Path-5 model surpasses Paprika by +0.28, +1.23, and +1.47 points on Task Recognition, Step Recognition, and Step Forecasting, respectively. Our slightly larger Path-6 model extends these gains even further. The improvements are even more pronounced with a simpler MLP head, where our method achieves gains of up to +2.24 points over Paprika. This confirms that representations pre-trained with our state-grounded, progressive method encode more salient and robust procedural knowledge, validating our overall approach.

\section{Conclusion}
In this work, we address a fundamental limitation in procedural video understanding: the failure of abstract task and step descriptions to align with concrete visual data. We introduce the TSS framework, a novel hierarchy that resolves this by grounding abstract procedures in observable, visually salient states. To teach this structure, we develop a progressive pre-training strategy that effectively unfolds the hierarchy, forcing the model to first ground its understanding in states before synthesizing knowledge of steps and tasks. Our state-of-the-art results on multiple benchmarks validate our central thesis: grounding abstract procedures in concrete, observable states is a crucial and previously missing component for robust procedural representation learning.

\section*{Acknowledgements}
This work was supported by Beijing Natural Science Foundation (L242019) and JSPS KAKENHI JP25K24384.

\bibliography{aaai2026}

\begin{thebibliography}{7}
\providecommand{\natexlab}[1]{#1}

\bibitem[Koupaee and Wang(2018)]{koupaee2018wikihow}
Koupaee, M.; and Wang, W.~Y. 2018.
\newblock Wikihow: A large scale text summarization dataset.
\newblock \textit{arXiv preprint arXiv:1810.09305}.

\bibitem[Miech et al.(2020)]{miech2020end}
Miech, A.; Alayrac, J.-B.; Smaira, L.; Laptev, I.; Sivic, J.; and Zisserman, A. 2020.
\newblock End-to-end learning of visual representations from uncurated instructional videos.
\newblock In \textit{Proceedings of the IEEE/CVF Conference on Computer Vision and Pattern Recognition}, 9879--9889.

\bibitem[Niu et al.(2024)]{niu2024schema}
Niu, Y.; Guo, W.; Chen, L.; Lin, X.; and Chang, S.-F. 2024.
\newblock Schema: State changes matter for procedure planning in instructional videos.
\newblock \textit{arXiv preprint arXiv:2403.01599}.

\bibitem[Samel, Sontakke, and Essa(2025)]{samel2025leveraging}
Samel, K.; Sontakke, N.; and Essa, I. 2025.
\newblock Leveraging procedural knowledge and task hierarchies for efficient instructional video pre-training.
\newblock \textit{arXiv preprint arXiv:2502.17352}.

\bibitem[Tang et al.(2019)]{tang2019coin}
Tang, Y.; Ding, D.; Rao, Y.; Zheng, Y.; Zhang, D.; Zhao, L.; Lu, J.; and Zhou, J. 2019.
\newblock Coin: A large-scale dataset for comprehensive instructional video analysis.
\newblock In \textit{Proceedings of the IEEE/CVF Conference on Computer Vision and Pattern Recognition}, 1207--1216.

\bibitem[Zhou et al.(2023)]{zhou2023procedure}
Zhou, H.; Mart{\'\i}n-Mart{\'\i}n, R.; Kapadia, M.; Savarese, S.; and Niebles, J.~C. 2023.
\newblock Procedure-aware pretraining for instructional video understanding.
\newblock In \textit{Proceedings of the IEEE/CVF Conference on Computer Vision and Pattern Recognition}, 10727--10738.

\bibitem[Zhukov et al.(2019)]{zhukov2019cross}
Zhukov, D.; Alayrac, J.-B.; Cinbis, R.~G.; Fouhey, D.; Laptev, I.; and Sivic, J. 2019.
\newblock Cross-task weakly supervised learning from instructional videos.
\newblock In \textit{Proceedings of the IEEE/CVF Conference on Computer Vision and Pattern Recognition}, 3537--3545.

\end{thebibliography}

\clearpage

\appendix

\twocolumn[{%
 \centering
 \LARGE \textbf{Appendix} \\[1.5em] 
}]

\setcounter{figure}{0}
\setcounter{table}{0}

\renewcommand{\thefigure}{A\arabic{figure}}
\renewcommand{\thetable}{A\arabic{table}}

\section{Data  Preprocessing}
To generate pseudo-labels, we first perform standardized preprocessing on the raw video and text data.



\subsection*{Video Preprocessing}
Following the procedures in \cite{zhou2023procedure,samel2025leveraging}, we segment videos in the temporal domain. Each video is divided into a series of non-overlapping segments, each with a duration of $T_S = 9.6\text{s}$. These 9.6s segments serve as the fundamental units for generating pseudo-labels at the task, step, and state levels. Prior to encoding, we preprocess the videos using \texttt{FFmpeg}: first, the frame rate is standardized to $F=10\text{ fps}$; second, each frame is center-cropped to a square aspect ratio and subsequently resized to a resolution of $224 \times 224$. Furthermore, each 9.6s segment is subdivided into 3 contiguous sub-clips, each with a duration of $t_s = 3.2\text{s}$. This results in a sequence of $t_s \times F = 32$ frames, which constitutes a single input for the visual encoder.

\subsection{State Description Generation}

To construct state-level pseudo-labels, we follow the methodology of Schema~\cite{niu2024schema} and leverage a Large Language Model (LLM) to generate detailed state descriptions. Specifically, we employ the \texttt{GPT-4o-mini} model for this task.

We utilize a prompt template for all 10,588 (task, step) pairs extracted from the WikiHow \cite{koupaee2018wikihow} dataset. This template, into which the specific `[task]' and `[step]' are dynamically inserted, instructs the model to infer the likely visual states corresponding to three distinct phases: \textit{before}, \textit{during}, and \textit{after} the execution of the given step within the context of the overarching task. The prompt template used to elicit these descriptions is detailed in Figure~\ref{fig:prompt}.

Figures~\ref{fig-state1}-\ref{fig-state4} showcase several representative examples of the generated state descriptions.

\subsection*{Feature Extraction}
We employ frozen pre-trained encoders to extract visual and textual features.

\paragraph{Visual Features} We use the S3D model provided by MIL-NCE \cite{miech2020end} as the visual encoder. For a given 9.6s segment $v$, its three sub-clips are encoded independently, yielding a feature matrix of shape $(3, 512)$. Subsequently, we apply average pooling across the first dimension to obtain a single vector $\boldsymbol{f}_v \in \mathbb{R}^{512}$ that represents the entire segment. Consequently, a video comprising $N$ segments is represented by a final visual feature matrix $\boldsymbol{f}_V \in \mathbb{R}^{N \times 512}$.

\paragraph{Textual Features}  
We encode task names, step headlines, and state descriptions with two text encoders: the S3D text module and Sentence-BERT’s MPNet (paraphrase-mpnet-base-v2). The resulting feature dimensions are 512-D for S3D and 768-D for MPNet. We perform clustering on the text features encoded by MPNet and generate pseudo-labels by matching them to the joint vision–text representations from S3D, using cosine distance as the similarity metric.

\begin{figure*}[t]          
\centering
\begin{tcolorbox}[myprompt]
\ttfamily\small
\kw{"role"}: "user",\\
\kw{"content"}: f"""
\par\smallskip
First, describe details of \meta{[step]} for \meta{[goal]} with one verb.
Second, use 3 sentences to describe status changes of objects before, transitioning, and after \meta{[step]}, avoiding using \meta{[word]}.
Additionally, apply common-sense constraints to the before/after states based on \meta{[last step]} and \meta{[next step]}.

\bigskip
\kw{[goal]}: Make Kimchi Fried Rice\\
\kw{[step]}: add ham\\
\kw{[word]}: add\\
\kw{Description}:\\
Add diced ham into the fried rice

\kw{Before}:
\begin{itemize}\setlength\itemsep{0pt}
  \item The diced ham is separate from the pan.
  \item The pan contains fried rice.
  \item The pan has no ham on it.
\end{itemize}

\kw{Transitioning}:
\begin{itemize}\setlength\itemsep{0pt}
  \item The diced ham is picked up from outside the pan \dots
  \item The diced ham is sprinkled into the pan \dots
  \item The diced ham gradually mixes \dots
\end{itemize}

\kw{After}:
\begin{itemize}\setlength\itemsep{0pt}
  \item The diced ham is mixed with the fried rice.
  \item The ham is on the pan.
  \item The pan contains ham.
\end{itemize}

\bigskip
\kw{[goal]}: \meta{\{task\}}\\
\kw{[step]}: \meta{\{step\}}\\
\kw{[word]}: \meta{\{step.split(' ')[0]\}}

\end{tcolorbox}
\caption{Prompt template for generating state descriptions.}
  \label{fig:prompt}
\end{figure*}

\section{Pseudo Label Generation}
After obtaining the video and text features, we proceed to generate pseudo-labels.

\subsubsection{Text Feature Clustering}

First, we perform text clustering to consolidate descriptions that are semantically similar but syntactically varied. This process aims to create a more compact and robust classification node space. Specifically, we employ bottom-up agglomerative clustering with cosine similarity as the distance metric. As each task is considered unique with its distinct sequence of steps, we only apply clustering to step-level and state-level texts. The clustering threshold is set to 0.09. To avoid confounding different temporal phases during state clustering, we cluster texts for ``before", ``mid", and ``after" states independently. Ultimately, this results in $N_{\text{task}}= 1053$ unique tasks (unclustered), ${N}_{\text{step}} = 10,038$ clustered step nodes, and the number of clustered state nodes (${N}_{\text{state}}$) being $9,755$ (${N}_{\text{state}}^b$), $9,289$ (${N}_{\text{state}}^m$), and $9,823$ (${N}_{\text{state}}^a$), respectively.

\subsubsection{Video-Node Matching (VNM) for Pseudo-Labels}

Next, we generate the TaskVNM, StepVNM, and StateVNM pseudo-labels by matching video clip features with the clustered text nodes. The matching logic is as follows:
For a given video clip feature, denoted as $f_v \in \mathbb{R}^{512}$, we compute its cosine similarity against the S3D-encoded features of all candidate texts. For StepVNM, this yields a similarity score vector covering all original step texts. Using the pre-established mapping from individual texts to their cluster nodes, these similarity scores are then aggregated (summed) for each unique node ID. The top 3 nodes with the highest aggregated scores are selected as the multi-class supervision target for StepVNM. For StateVNM, we first determine the state's temporal type (before/mid/after) by identifying the type of the best-matching raw state description for $f_v$. Subsequently, the node matching and score aggregation process is performed exclusively within the pool of nodes belonging to that determined type.

\subsubsection{Contextual and Relational Pseudo-Labels (TCL \& NRL)}

For StepTCL and StepNRL, the generation process follows the methodology from \cite{zhou2023procedure}, briefly described as follows:
Both StepTCL and StepNRL are built upon the foundation of StepVNM. StepTCL aims to identify other step nodes that are contextually relevant within the same task as the step node associated with the current video clip. We first construct a task-step co-occurrence matrix (shape: $N_\text{task} \times N_\text{step}$) based on procedural data from WikiHow. Then, using TaskVNM, we identify the task associated with the video clip. With the co-occurrence matrix, we find all steps related to this task, aggregate the scores onto their corresponding node IDs, and select the top 3 as the supervision target.

StepNRL is designed to learn the sequential transition patterns between steps. This transition information is derived from two sources: 
\begin{itemize}
    \item[1)] The prescribed step sequences within each task from WikiHow.
    \item[2)] The inherent chronological sequence of video clips in HowTo100M.
\end{itemize}
By defining a ``hop" distance, we can identify ``hop-in" (nodes from which a directed path of length `hop' exists to the current node) and ``hop-out" (nodes to which a directed path of length `hop' exists from the current node) lists for any given node. Due to computational constraints, we set the hop distance to 1.

For all pseudo-label types mentioned above, we select the top 3 scoring nodes as the supervision targets for model pre-training.

Figure \ref{fig-pseduolabel} shows some of the pseudo-labels we obtained.
\begin{figure}[htbp]
\centering
\includegraphics[width=\linewidth, page=1]{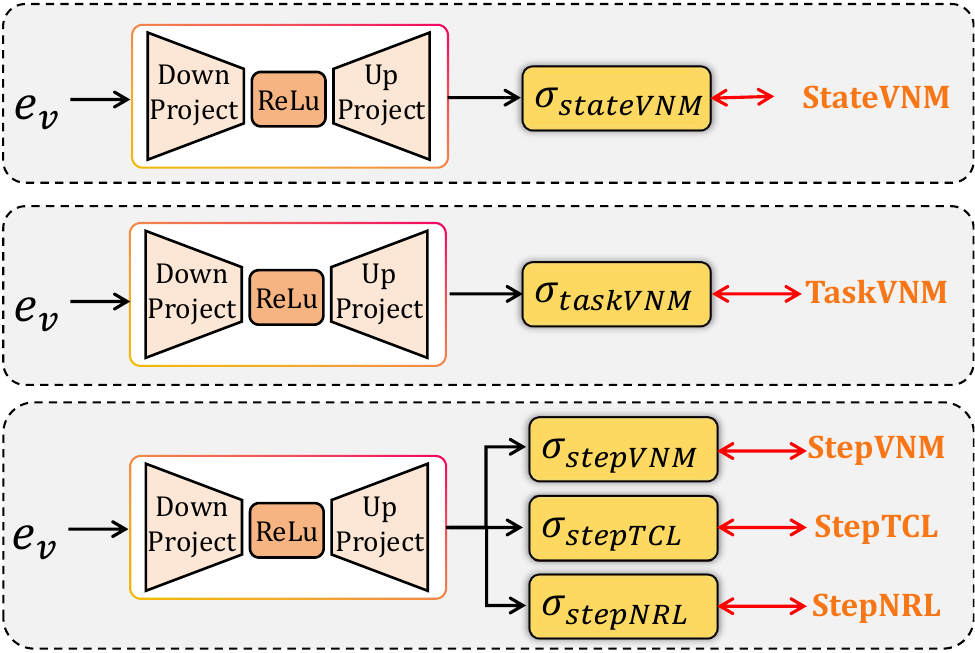}
\caption{Pre-
training process for three layers of the
TSS framework.}
\label{fig-forward}
\end{figure}

\begin{figure*}[htbp]
\centering
\includegraphics[width=\linewidth, page=1]{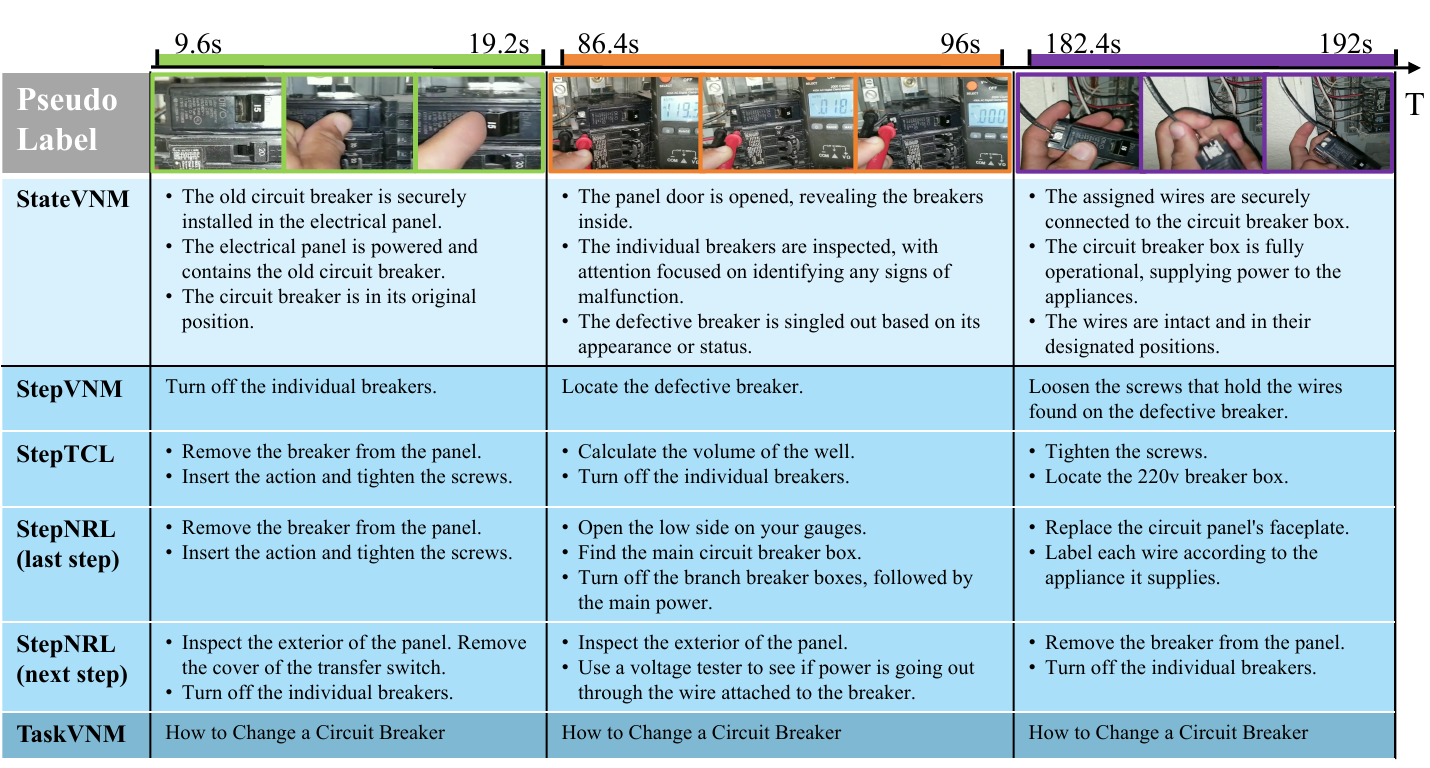}
\caption{Example of pseudo label.
}
\label{fig-pseduolabel}
\end{figure*}

\section{Progressive Training Setting}

During pre-training, two main trainable modules are involved: the \textbf{adapter}, denoted as $\phi_2$, and the \textbf{task-specific head}, denoted as $\sigma$.

The \textbf{adapter} ($\phi_2$) employs a bottleneck architecture. It consists of two linear layers with an intermediate ReLU activation function, which projects the 512-dimensional features down to 128 dimensions and then maps them back to the original 512 dimensions.

The \textbf{task-specific head} ($\sigma$)is a Multi-Layer Perceptron (MLP) designed to predict the pseudo-labels. Its structure comprises three linear layers, with dimensional transformations following $d_{in} \to d_{h1} \to d_{h2} \to d_{out}$. We set the hidden dimensions as $d_{h1} = d_{out} // 4$ and $d_{h2} = d_{out} // 2$. Notably, for the StepNRL task (with a hop distance of 1), we decompose it into two sub-tasks, StepNRL-in and StepNRL-out, and consequently utilize two separate task heads.

Our progressive pre-training strategy is composed of five stages. Each stage focuses on a specific layer within the TSS framework, utilizing the corresponding pseudo-labels and a dedicated task head for training. Figure \ref{fig-forward} illustrates this pre-training process for the three layers of the TSS framework.

The pre-training process was implemented using the \textbf{PyTorch} framework. For the multi-classification task, we employed the \textbf{Adam} optimizer and used Binary Cross-Entropy Loss with Logits (\texttt{BCEWithLogitsLoss}) as the loss function. The hyperparameters were configured as follows: a learning rate of $1 \times 10^{-4}$, a weight decay of 0, and a batch size of 256. The training set consisted of 4.1 million video clip samples.

Training was conducted on a server equipped with 8 H200 GPUs and ran for 1500 epochs. The duration for a single epoch was approximately 90 seconds during the Step or State phase and approximately 30 seconds during the Task phase.

\section{downstream eval settings}
Downstream task evaluation was performed on two datasets: \textbf{COIN} \cite{tang2019coin} and \textbf{CrossTask} \cite{zhukov2019cross}.

The \textbf{COIN} dataset comprises 11,827 videos across 180 tasks, with precise temporal annotations provided for 749 distinct steps.

The \textbf{CrossTask} dataset contains 83 tasks, uniquely divided into 18 ``primary" tasks with full temporal step annotations and 65 ``related" tasks without them. This dataset includes annotations for 105 different steps.

Downstream tasks were evaluated using two model architectures: a \textbf{Multi-Layer Perceptron (MLP)} and a \textbf{Transformer}.

\begin{itemize}
    \item \textbf{MLP}: The MLP structure consisted of a positional encoding layer, a ReLU activation layer, and two linear layers with dimensions $d_{\text{input}} \rightarrow d_{\text{hidden}} \rightarrow d_{\text{output}}$. The hidden dimension ($d_{\text{hidden}}$) was set to 768 for Step Recognition (SR) and Step Forecasting (SF), and to 128 for Task Recognition (TR).
    \item \textbf{Transformer}: For the Transformer model, a Transformer encoder layer was added before the MLP. The temporal feature sequence was first augmented with absolute positional encoding information and then prepended with a learnable \texttt{[CLS]} token. The output of this \texttt{[CLS]} token served as the input to the MLP. The Transformer encoder layer utilized 8 attention heads and a hidden dimension of 1024.
\end{itemize}

The hyperparameters for downstream evaluation were set as follows: a batch size of 16, a learning rate of $1 \times 10^{-4}$, a weight decay of $1 \times 10^{-3}$, and the \textbf{Adam} optimizer. Early stopping patience was set to 50.

\begin{figure*}[htbp]
\centering
\includegraphics[width=0.8\linewidth, page=1]{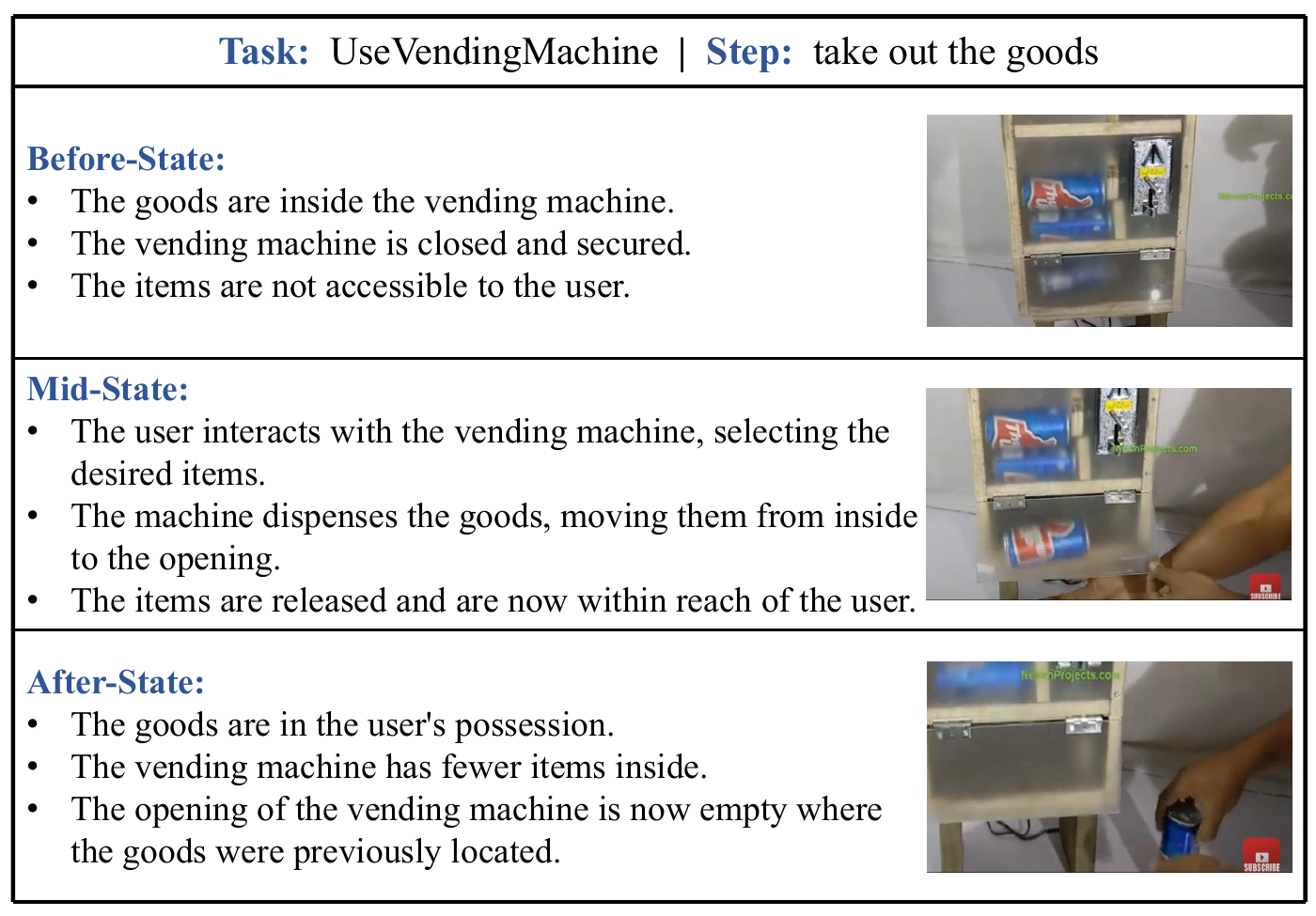}
\caption{Example of partial state description (1)}
\label{fig-state1}
\end{figure*}

\begin{figure*}[htbp]
\centering
\includegraphics[width=0.82\linewidth, page=2]{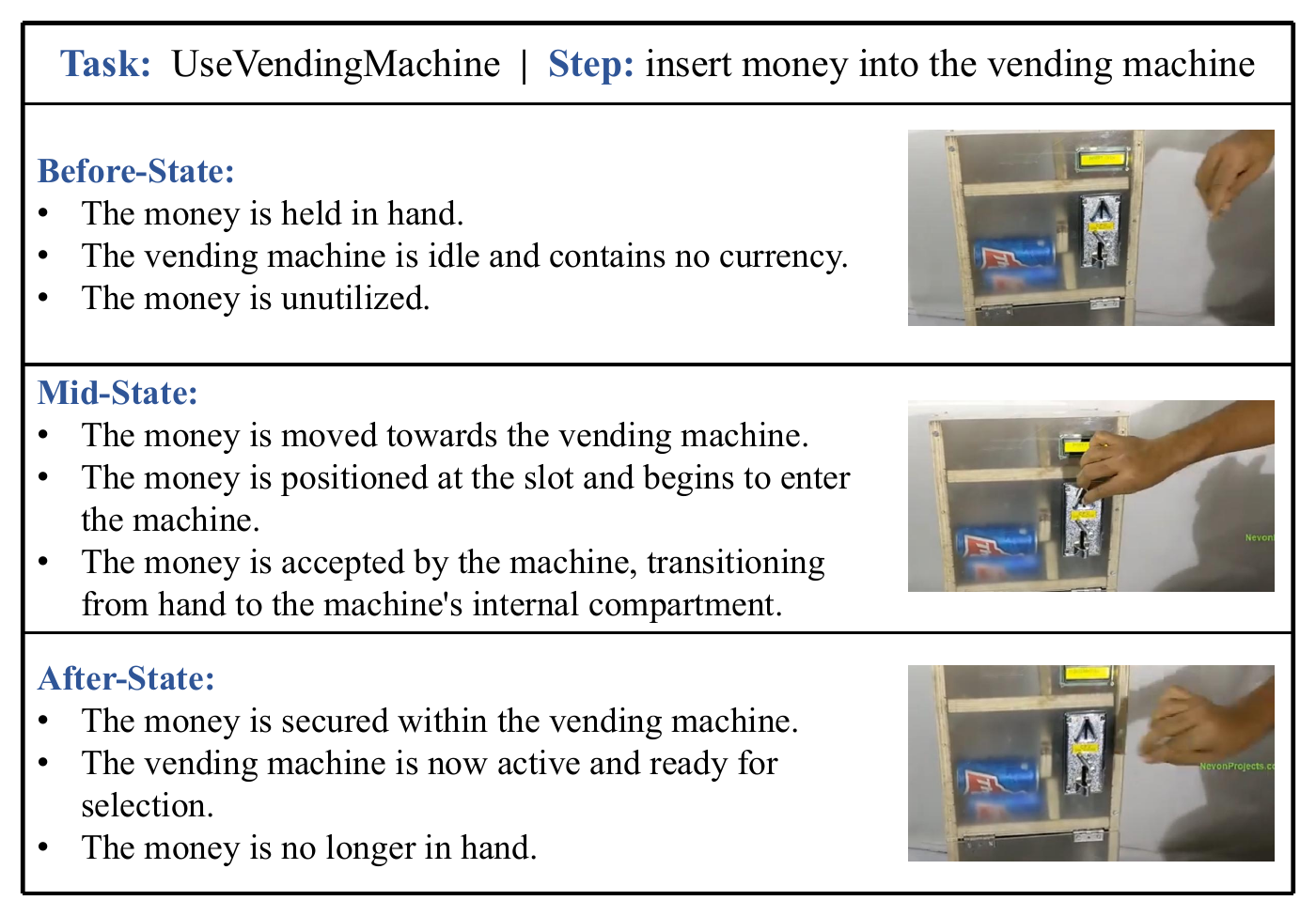}
\caption{Example of partial state description (2)}
\label{fig-state2}
\end{figure*}

\begin{figure*}[htbp]
\centering
\includegraphics[width=0.8\linewidth, page=3]{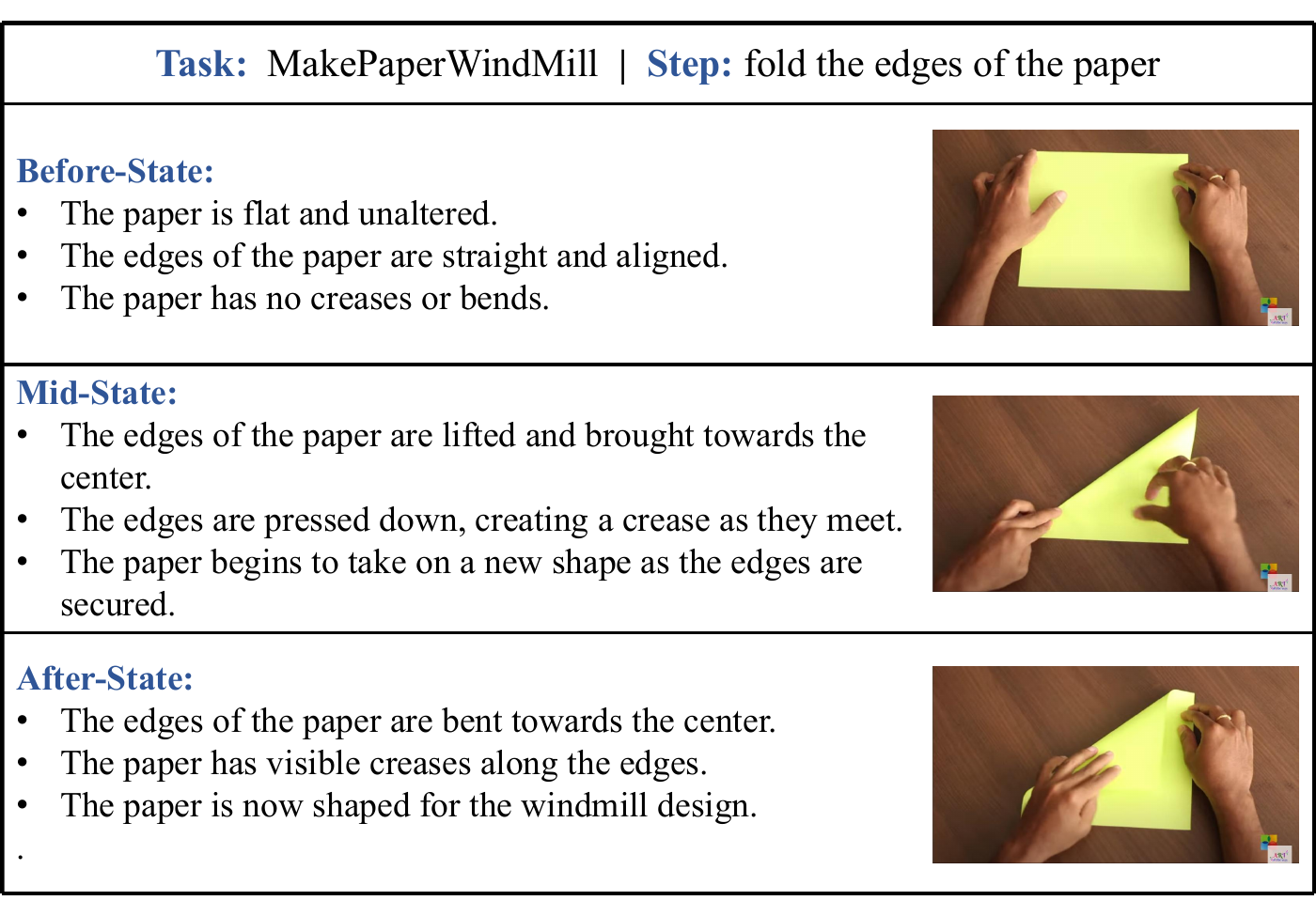}
\caption{Example of state description (3) }
\label{fig-state3}
\end{figure*}

\begin{figure*}[htbp]
\centering
\includegraphics[width=0.85\linewidth, page=4]{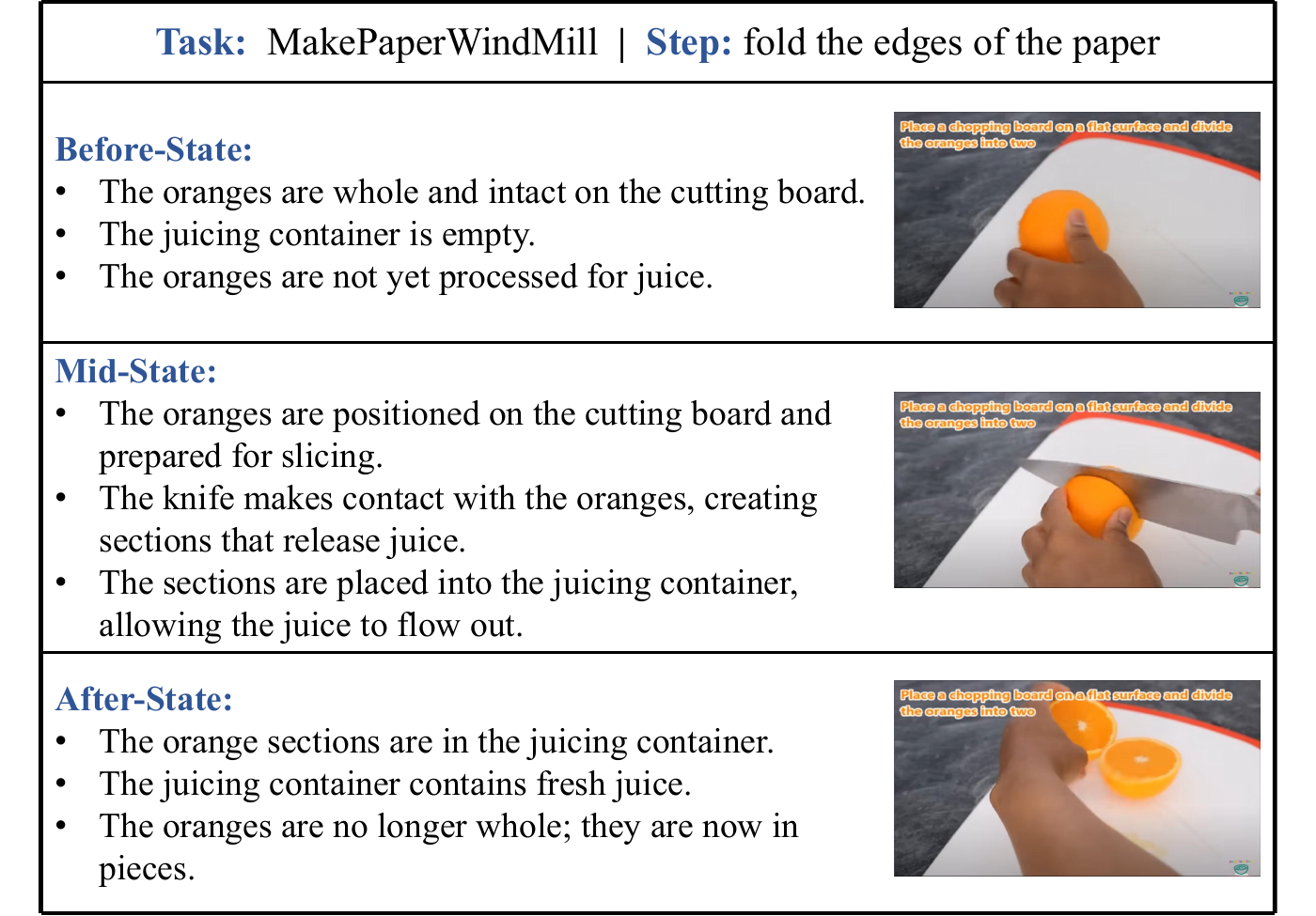}
\caption{Example of state description (4) }
\label{fig-state4}
\end{figure*}


\end{document}